\documentclass[aos,preprint]{imsart}
\setattribute{journal}{name}{} 

\RequirePackage[OT1]{fontenc}
\RequirePackage{amsthm,amsmath,amssymb}

\RequirePackage{natbib}
\setcitestyle{round,authoryear,citesep={;},aysep={,},yysep={;}}

\renewcommand{\cite}[1]{\citep{#1}}

\RequirePackage[colorlinks,citecolor=blue,urlcolor=blue]{hyperref}

\usepackage{graphicx} 

\usepackage{macros}

\usepackage{caption}
\usepackage{subcaption}

\usepackage{verbatim}
\usepackage{gensymb}

\usepackage[top=1in, bottom=1in, left=1in, right=1in]{geometry}

\begin{document}

\begin{frontmatter}
\title{Discovering Latent Network Structure in \\Point Process Data}
\runtitle{Discovering Latent Network Structure in Point Process Data}

\begin{aug}
\author{\fnms{Scott W.} \snm{Linderman}\ead[label=e1]{slinderman@seas.harvard.edu}}
\and
\author{\fnms{Ryan P.} \snm{Adams}\ead[label=e2]{rpa@seas.harvard.edu}}
\affiliation{Harvard University}

\runauthor{S. W. Linderman and R. P. Adams}
\end{aug}

\begin{abstract}
Networks play a central role in modern data analysis, enabling us to reason about systems by studying the relationships between their parts.  Most often in network analysis, the edges are given.  However, in many systems it is difficult or impossible to measure the network directly.  Examples of latent networks include economic interactions linking financial instruments and patterns of reciprocity in gang violence.  In these cases, we are limited to noisy observations of events associated with each node.  To enable analysis of these implicit networks, we develop a probabilistic model that combines mutually-exciting point processes with random graph models.  We show how the Poisson superposition principle enables an elegant auxiliary variable formulation and a fully-Bayesian, parallel inference algorithm.  We evaluate this new model empirically on several datasets.
\end{abstract}
\end{frontmatter}

\section{Introduction}
Many types of modern data are characterized via relationships on a network.  Social network analysis is the most commonly considered example, where the properties of individuals (vertices) can be inferred from ``friendship'' type connections (edges).  Such analyses are also critical to understanding regulatory biological pathways, trade relationships between nations, and propagation of disease.  The tasks associated with such data may be unsupervised (e.g., identifying low-dimensional representations of edges or vertices) or supervised (e.g., predicting unobserved links in the graph).  Traditionally, network analysis has focused on \emph{explicit network} problems in which the graph itself is considered to be the observed data.  That is, the vertices are considered known and the data are the entries in the associated adjacency matrix. A rich literature has arisen in recent years for applying statistical machine learning models to this type of problem, e.g., \citet{Liben-2007,Hoff-2008,Goldenberg-2010}.

In this paper we are concerned with \emph{implicit networks} that cannot be observed directly, but about which we wish to perform analysis.  In an implicit network, the vertices or edges of the graph may not be directly observed, but the graph structure may be inferred from noisy emissions.  These noisy observations are assumed to have been generated according to underlying dynamics that respect the latent network structure.

For example, trades on financial stock markets are executed thousands of times per second. Trades of one stock are likely to cause subsequent activity on stocks in related industries. How can we infer such interactions and disentangle them from market-wide fluctuations that occur throughout the day? Discovering latent structure underlying financial markets not only reveals interpretable patterns of interaction, but also provides insight into the stability of the market. In Section~\ref{sec:stability} we will analyze the stability of mutually-excitatory systems, and in Section~\ref{sec:financial} we will explore how stock similarity may be inferred from trading activity.

As another example, both the edges and vertices may be latent.  In Section~\ref{sec:chicago}, we examine patterns of violence in Chicago, which can often be attributed to social structures in the form of gangs.  We would expect that attacks from one gang onto another might induce cascades of violence, but the vertices (gang identity of both perpetrator and victim) are unobserved.  As with the financial data, it should be possible to exploit dynamics to infer these social structures.  In this case spatial information is available as well, which can help inform latent vertex identities.

In both of these examples, the noisy emissions have the form of events in time, or ``spikes,'' and our intuition is that a spike at a vertex will induce activity at adjacent vertices.  In this paper, we formalize this idea into a probabilistic model based on mutually-interacting point processes.  Specifically, we combine the Hawkes process \cite{Hawkes-1971} with recently developed exchangeable random graph priors.  This combination allows us to reason about latent networks in terms of the way that they regulate interaction in the Hawkes process.  Inference in the resulting model can be done with Markov chain Monte Carlo, and an elegant data augmentation scheme results in efficient parallelism.

\section{Preliminaries}
\subsection{Poisson Processes}
Point processes are fundamental statistical objects that yield random finite sets of events~${\{s_n\}_{n=1}^N \subset \mathcal{S}}$, where~$\mathcal{S}$ is a compact subset of~${\reals^D}$, for example, space or time. The Poisson process is the canonical example. It is governed by a nonnegative ``rate'' or  ``intensity'' function,~${\lambda(s): \mathcal{S}\rightarrow\reals_+}$. The number of events in a subset~${\mathcal{S}'\subset\mathcal{S}}$ follows a Poisson distribution with mean~${\int_{\mathcal{S}'}\lambda(s)\mathrm{d}s}$. Moreover, the number of events in disjoint subsets are independent. 

We use the notation~${\{s_n\}_{n=1}^N\sim\PP(\lambda(s))}$ to indicate that a set of events~$\{s_n\}_{n=1}^N$ is drawn from a Poisson process with rate~$\lambda(s)$. The likelihood is given by
\begin{align}
\label{eq:poisson_lkhd}
p(\{s_n\}_{n=1}^N|\lambda(s))=\exp\left\{-\!\int_{\mathcal{S}}\!\lambda(s)\mathrm{d}s\right\}\prod_{n=1}^N\lambda(s_n).
\end{align}

In this work we will make use of a special property of Poisson processes, the \emph{Poisson superposition theorem}, which states that ~${\{s_n\}\sim\PP(\lambda_1(s)+\ldots+\lambda_K(s))}$ can be decomposed into~$K$ independent Poisson processes. Letting~${z_n}$ denote the origin of the~$n$-th event, we perform the decomposition by independently sampling each~${z_n}$ from~${\Pr(z_n=k)\propto\lambda_k(s_n)}$, for~${k\in\{1\ldots K\}}$ \cite{Daley-1988}. 

\subsection{Hawkes Processes}
Though Poisson processes have many nice properties, they cannot capture interactions between events. For this we turn to a more general model known as Hawkes processes. A Hawkes process consists of~$K$ point processes and gives rise to sets of \emph{marked} events $\{s_n,c_n\}_{n=1}^N$, where~${c_n\in\{1,\ldots,K\}}$ specifies the process on which the~$n$-th event occurred. For now, we assume the events are points in time, i.e.,~${s_n\in[0,T]}$. 
Each of the~$K$ processes is a \emph{conditionally Poisson process} with a rate~${\lambda_k(t\given \{s_n:s_n<t\})}$ that depends on the history of events up to time~$t$.

Hawkes processes have additive interactions. Each process has a ``background rate'' $\lambda_{0,k}(t)$, and each event~$s_n$ on process $k$ adds a nonnegative impulse response~$h_{k,k'}(t-s_n)$ to the intensity of other processes~$k'$. Causality and locality of influence are enforced by requiring~$h_{k,k'}(\Delta t)$ to be zero for~${\Delta t \notin[0,\Delta t_{\mathsf{max}}]}$. 
\begin{figure}[t]
\begin{center}
\includegraphics[width=.66\linewidth]{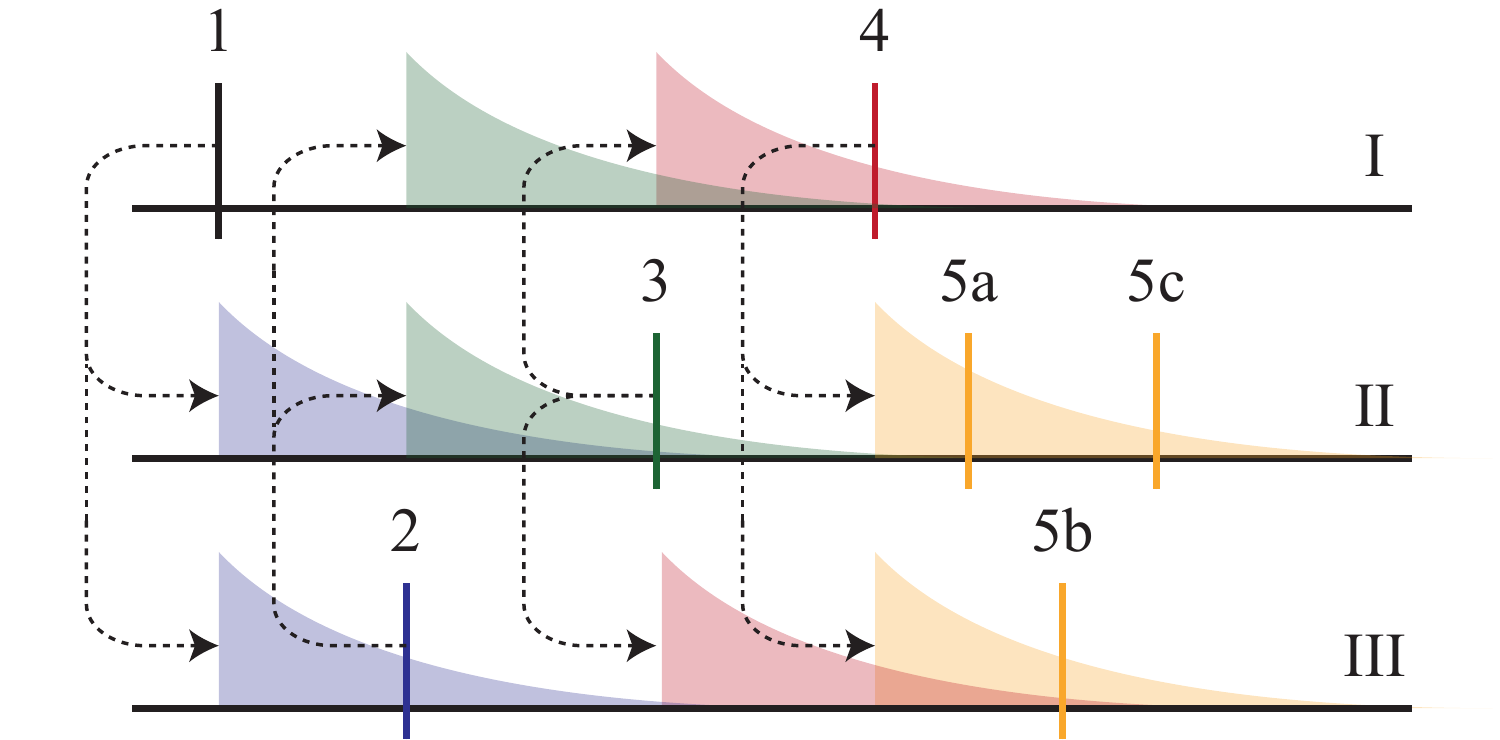} 
\end{center}
\caption{Illustration of a Hawkes process. Events induce impulse responses on connected processes and spawn ``child'' events. See the main text for a complete description.}
\label{fig:hawkes}
\end{figure}

By the superposition theorem for Poisson processes, these additive components can be considered independent processes, each giving rise to their own events. 
We augment our data with a latent random variable~${z_n\in\{0,\ldots,n-1\}}$ to indicate the cause of the~$n$-th event ($0$ if the event is due to the background rate and ${1\ldots n-1}$ if it was caused by a preceding event). 

Let~$\mathcal{C}_{n,k'}$ denote the set of events on process~$k'$ that were parented by event $n$. Formally,
\begin{align*}
  \mathcal{C}_{n,k'} \equiv \{s_{n'}: c_{n'}=k' \wedge z_{n'}=n\}.
\end{align*}
Let $\mathcal{C}_{0,k}$ be the set of events attributed to the background rate of process $k$.  The augmented Hawkes likelihood is the product of likelihoods of each Poisson process:
\begin{align}
\nonumber p(\{(s_n,c_n,z_n)\}^N_{n=1} \given \{\lambda_{0,k}(t)\},& \{\{h_{k,k'}(\Delta t)\}\}) = \\
&\qquad\left[\prod^K_{k=1} p(\mathcal{C}_{0,k}\given \lambda_{0,k}(t))\right]\times\left[\prod_{n=1}^N\prod_{k=1}^K p(\mathcal{C}_{n,k}\given h_{c_n,k}(t-s_n))\right],
\label{eq:hawkes_likelihood}
\end{align}
where the densities in the product are given by Equation~\ref{eq:poisson_lkhd}.

Figure~\ref{fig:hawkes} illustrates a causal cascades of events for a simple network of three processes (I-III).  The first event is caused by the background rate~(${z_1=0}$), and it induces impulse responses on processes II and III. Event~2 is spawned by the impulse on the third process~(${z_2=1}$), and feeds back onto processes I and II. In some cases a single parent event induces multiple children, e.g., event~4 spawns events~{5a-c}. In this simple example, processes excite one another, but do not excite themselves. Next we will introduce more sophisticated models for such interaction networks.  

\subsection{Random Graph Models}
\label{sec:graph_models}
Graphs of~$K$ nodes correspond to~${K\times K}$ matrices. Unweighted graphs are binary adjacency matrices~$\bA$ where~${A_{k,k'}=1}$ indicates a directed edge from node~$k$ to node~$k'$. Weighted directed graphs can be represented by a real matrix $\bW$ whose entries indicate the weights of the edges. Random graph models reflect the probability of different network structures through distributions over these matrices.

Recently, many random graph models have been unified under an elegant theoretical framework due to Aldous and Hoover \cite{Aldous-1981,Hoover-1979}. See \citet{Lloyd-2012} for an overview. Conceptually, the Aldous-Hoover representation characterizes the class of \textit{exchangeable} random graphs, that is, graph models for which the joint probability is invariant under permutations of the node labels. Just as de Finetti's theorem equates exchangeable sequences  $(X_n)_{n\in\mathbb{N}}$ to independent draws from a random probability measure $\Theta$, the Aldous-Hoover theorem relates random exchangeable graphs to the following generative model:
\begin{align*}
  u_1,u_2,\ldots&\sim_{i.i.d}\text{Uniform}[0,1],\\
  A_{k,k'}&\sim \text{Bernoulli}(\Theta(u_k,u_{k'})),
\end{align*}
for some random function $\Theta:[0,1]^2\rightarrow [0,1]$.

Empty graph models~(${A_{k,k'}\equiv 0}$) and complete models~(${A_{k,k'}\equiv 1}$) are trivial examples, but much more structure may be encoded. For example, consider a model in which nodes are endowed with a location in space,~${\bx_k\in\reals^D}$. This could be an abstract feature space or a real location like the center of a gang territory. The probability of connection between two notes decreases with distance between them as~${A_{k,k'}\sim\text{Bern}(\rho e^{-||\bx_k-\bx_{k'}||/\tau})}$, where~$\rho$ is the overall sparsity and~$\tau$ is the characteristic distance scale. This simple model can be converted to the Aldous-Hoover representation by transforming~$u_k$ into~${\bx_k}$ via the inverse CDF. 

Many models can be constructed in this manner. Stochastic block models, latent eigenmodels, and their nonparametric extensions all fall under this class \cite{Lloyd-2012}. We will leverage the generality of the Aldous-Hoover formalism to build a flexible model and inference algorithm for Hawkes processes with structured interaction networks.

\section{The Network Hawkes Model}\label{sec:basic_model}
In order to combine Hawkes processes and random network models, we decompose the Hawkes impulse response $h_{k,k'}(\Delta t)$ as follows:
\begin{align}
\label{eq:ir_decomp}
h_{k,k'}(\Delta t)=A_{k,k'}W_{k,k'}g_{\theta_{k,k'}}(\Delta t).
\end{align}
Here, ${\bA\in\{0,1\}^{K\times K}}$ is a binary adjacency matrix and~${\bW\in\reals_+^{K\times K}}$ is a non-negative weight matrix. Together these specify the \emph{sparsity structure} and \emph{strength} of the interaction network, respectively. The non-negative function~${g_{\theta_{k,k'}}(\Delta t)}$ captures the temporal aspect of the interaction. It is parameterized by~${\theta_{k,k'}}$ and satisfies two properties: a) it has bounded support for~${\Delta t \in [0,\Delta t_{\mathsf{max}}]}$, and~b) it integrates to one. In other words,~$g$ is a probability density with compact support.

Decomposing~$h$ as in Equation~\ref{eq:ir_decomp} has many advantages. It allows us to express our separate beliefs about the sparsity structure of the interaction network and the strength of the interactions through a spike-and-slab prior on~$\bA$ and~$\bW$ \cite{Mohamed-2012}. 
The empty graph model recovers independent background processes, and the complete graph recovers the standard Hawkes process.
Making~$g$ a probability density endows~$\bW$ with units of ``expected number of events'' and allows us to compare the relative strength of interactions. The form suggests an intuitive generative model: for each impulse response draw~${m\sim \text{Poisson}(W_{k,k'})}$ number of induced events and draw the~$m$ child event times i.i.d.\ from~$g$, enabling computationally tractable conjugate priors.

Intuitively, the background rates,~$\lambda_{0,k}(t)$, explain events that cannot be attributed to preceding events. In the simplest case the background rate is constant. However, there are often fluctuations in overall intensity that are shared among the processes, and not reflective of process-to-process interaction, as we will see in the daily variations in trading volume on the S\&P100 and the seasonal trends in homicide. To capture these shared background fluctuations, we use a sparse Log Gaussian Cox process \cite{Moller-1998} to model the background rate:
\begin{align*}
  \lambda_{0,k}(t)=\mu_{k} + \alpha_{k}\exp\{\by(t)\},\;\;\by(t)\sim\mathcal{GP}(\boldsymbol{0},K(t,t')).
\end{align*}

The kernel~${K(t,t')}$ describes the covariance structure of the background rate that is shared by all processes. For example, a periodic kernel may capture seasonal or daily fluctuations. The offset~${\mu_k}$ accounts for varying background intensities among processes, and the scaling factor~$\alpha_k$ governs how sensitive process~$k$ is to these background fluctuations (when~${\alpha_k=0}$ we recover the constant background rate).


Finally, in some cases the process identities,~${c_n}$, must also be inferred. With gang incidents in Chicago we may have only a location,~${\bx_n\in\reals^2}$. In this case, we may place a spatial Gaussian mixture model over the~$c_n$'s, as in \citet{Cho-2013}. Alternatively, we may be given the label of the community in which the incident occurred, but we suspect that interactions occur between clusters of communities. In this case we can use a simple clustering model or a nonparametric model like that of \citet{Blundell-2012}. 

\subsection{Inference with Gibbs Sampling}
We present a Gibbs sampling procedure for inferring the model parameters,~$\bW$,~$\bA$,~$\{\{\theta_{k,k'}\}\}$,$\{\lambda_{0,k}(t)\}$, and, if necessary,~${\{c_n\}}$. In order to simplify our Gibbs updates, we will also sample a set of parent assignments for each event~$\{z_n\}$. Incorporating these parent variables enables conjugate prior distributions for~$\bW$,~$\theta_{k,k'}$, and, in the case of constant background rates,~$\lambda_{0,k}$.

\paragraph{Sampling weights $\bW$.} A gamma prior on the weights, ${W_{k,k'}\sim\distGamma(\alpha_W^{0}, \beta_W^{0})}$, results in the conditional distribution,
\begin{align*}
&W_{k,k'}\given \{s_n,c_n,z_n\}^N_{n=1}, \theta_{k,k'} \sim
\distGamma(\alpha_{k,k'}, \beta_{k,k'}),\\ 
&\qquad\alpha_{k,k'} = \alpha_W^0 + \sum_{n=1}^N\sum_{n'=1}^N \delta_{c_n,k}\delta_{c_{n'},k'}\delta_{z_{n',}n}\\ 
&\qquad\beta_{k,k'} = \beta_W^0 + \sum_{n=1}^N \delta_{c_n,k}.
\end{align*}
This is a minor approximation valid for~$\Delta t_{\mathsf{max}} \ll T$.  Here and elsewhere, $\delta_{i,j}$ is the Kronecker delta function. We use the inverse-scale parameterization of the gamma distribution, i.e.,
\begin{align*}
\distGamma(x\given\alpha,\beta) = \frac{\beta^\alpha}{\Gamma(\alpha)} x^{\alpha-1} \exp\{-\beta\,x\}.
\end{align*}

\paragraph{Sampling impulse response parameters $\theta_{k,k'}$.} We let~$g_{k,k'}(\Delta t)$ be the logistic-normal density with parameters~$\theta_{k,k'}=\{\mu,\tau\}$: 
\begin{align*}
g_{k,k'}(\Delta t\given \mu,\tau)&=\frac{1}{Z}\exp\left\{\frac{-\tau}{2}\left(\sigma^{-1}\left(\frac{\Delta t}{\Delta t_{\mathsf{max}}}\right) - \mu\right)^2\right\}\\
\sigma^{-1}(x) &= \ln(x/(1-x))\\
Z &= \frac{\Delta t(\Delta t_{\sf{max}}-\Delta t)}{\Delta t_{\mathsf{max}}} \left(\frac{\tau}{2\pi}\right)^{-\frac{1}{2}}.
\end{align*}
The normal-gamma prior 
\begin{align*}
\mu,\tau &\sim \mathcal{NG}(\mu,\tau|\mu_\mu^0,\kappa_\mu^0,\alpha_\tau^0,\beta_\tau^0)
\end{align*}
yields the standard conditional distribution \citep[see][]{Murphy-2012} with the following sufficient statistics:
\begin{align*}
&\quad x_{n,n'} = \ln(s_{n'}-s_n)-\ln(t_{\sf{max}}-(s_{n'}-s_n)), \\
&\quad m = \sum_{n=1}^N\sum_{n'=1}^N \delta_{c_n,k}\delta_{c_{n'},k'}\delta_{z_{n'},n}, \\
&\quad\bar{x} = \frac{1}{m} \sum_{n=1}^N\sum_{n'=1}^N \delta_{c_n,k}\delta_{c_{n'},k'}\delta_{z_{n'},n}x_{n,n'}.
\end{align*}

\paragraph{Sampling background rates $\lambda_{0,k}$.} For background rates~$\lambda_{0,k}(t)\equiv\lambda_{0,k}$, the prior~${\lambda_{0,k} \sim \distGamma(\alpha_{\lambda}^0,\beta_{\lambda}^0)}$ is conjugate with the likelihood and yield the conditional distribution 
\begin{align*}
  &\lambda_{0,k}\given \{s_n,c_n,z_n\}^N_{n=1},\sim \distGamma(\alpha_{\lambda},\;  \beta_\lambda),\\
  &\qquad \alpha_{\lambda} = \alpha^0_{\lambda} + \sum_n \delta_{c_n,k}\delta_{z_n,0}\\
  &\qquad \beta_\lambda = \beta_\lambda^0 + T
\end{align*}

This conjugacy no longer holds for Gaussian process background rates, but conditioned upon the parent variables, we must simply fit a Gaussian process for those events for which~${z_n=0}$. We use elliptical slice sampling \cite{Murray-2010} for this purpose. 

\paragraph{Collapsed Gibbs sampling $\bA$ and $z_n$.}
With Aldous-Hoover graph priors, the entries in the binary adjacency matrix~$\bA$ are conditionally independent given the parameters of the prior. The likelihood introduces dependencies between the rows of~$\bA$, but each column can be sampled in parallel. Gibbs updates are complicated by strong dependencies between the graph and the parent variables,~$z_n$. Specifically, if~${z_{n'}=n}$, then we must have~${A_{c_{n},c_{n'}}=1}$. To improve the mixing of our sampling algorithm, first we update~${\bA\given\{s_n,c_n\},\bW,\theta_{k,k'}}$ by marginalizing the parent variables. The posterior is determined by the likelihood of the conditionally Poisson process~${\lambda_{k'}(t\given\{s_n:s_n<t\})}$ (Equation~\ref{eq:poisson_lkhd}) with and without interaction~${A_{k,k'}}$ and the prior comes from the Aldous-Hoover graph model. Then we update~${z_n\given\{s_n,c_n\},\bA,\bW,\theta_{k,k'}}$ by sampling from the discrete conditional distribution. Though there are~$N$ parent variables, they are conditionally independent and may be sampled in parallel. We have implemented our inference algorithm on GPUs to capitalize on this parallelism.

\paragraph{Sampling process identities $c_n$.}
As with the adjacency matrix, we use a collapsed Gibbs sampler to marginalize out the parent variables when sampling the process identities. Unfortunately, the~$c_n$'s are not conditionally independent and hence must be sampled sequentially. This limits the size of the datasets we can handle when the process identities are unknown, but our GPU implementation is still able to achieve upwards of 4 iterations (sampling all variables) per second on datasets with thousands of events.

\section{Stability of Network Hawkes Processes}
\label{sec:stability}
\begin{figure*}[t]
\vspace{-.5em}
\begin{center}
\begin{subfigure}[b]{.23\textwidth}
\caption{}
\label{fig:stability_p_w}
\vspace{-.9em}
\includegraphics[height=1.45in]{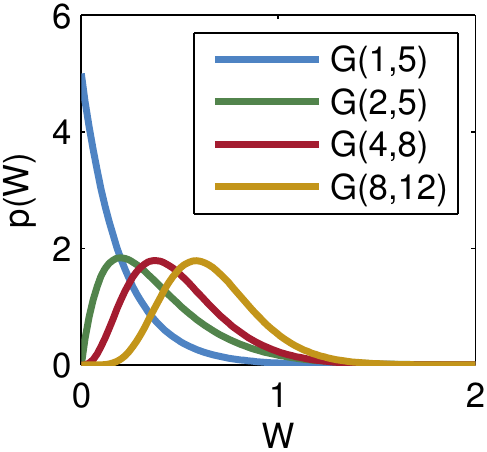} 
\end{subfigure}
~
\begin{subfigure}[b]{.23\textwidth}
\caption{}
\vspace{-1em}
\label{fig:stability_max_rho}
\includegraphics[height=1.45in]{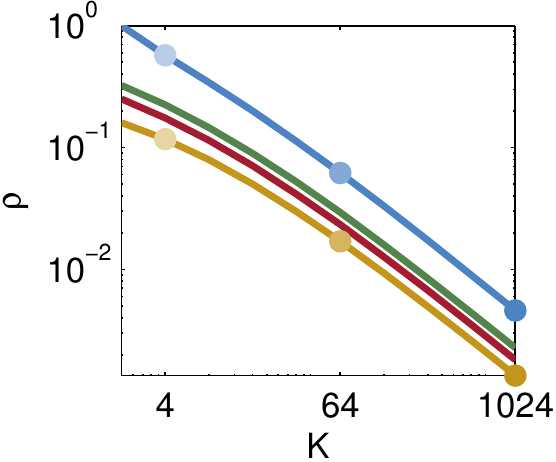} 
\end{subfigure}
~
\hspace{1em}
\begin{subfigure}[b]{.23\textwidth}
\caption{}
\label{fig:stability_1_5}
\vspace{-.5em}
\includegraphics[height=1.4in]{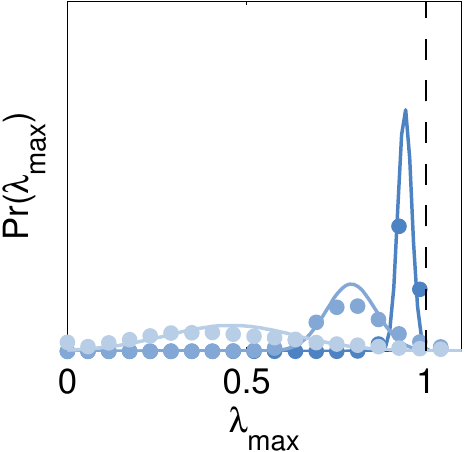} 
\end{subfigure}
~
\begin{subfigure}[b]{.23\textwidth}
\caption{}
\label{fig:stability_8_12}
\vspace{-.5em}
\includegraphics[height=1.4in]{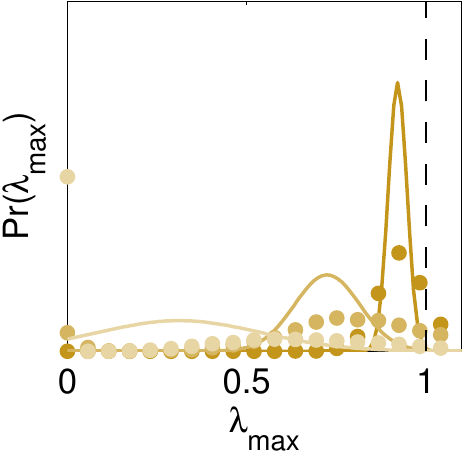} 
\end{subfigure}
\end{center}
\vspace{-1em}
\caption{Empirical and theoretical distribution of the maximum eigenvalue for Erd\H{o}s-Renyi graphs with gamma weights. (a) Four gamma weight distributions. The colors correspond to the curves in the remaining panels. (b) Sparsity that theoretically yields ${99\%}$ probability of stability as a function of~${p(W)}$ and~$K$. (c) and (d) Theoretical (solid) and empirical (dots) distribution of the maximum eigenvalue. Color corresponds to the weight distribution in (a) and intensity indicates~$K$ and~$\rho$ shown in (b).}
\label{fig:stability}
\end{figure*}

Due to their recurrent nature, Hawkes processes must be constrained to ensure their positive feedback does not lead to infinite numbers of events. A stable system must satisfy\footnote{In this context~${\lambda_{\mathsf{max}}}$ refers to an eigenvalue rather than a rate, and~$\odot$ denotes the Hadamard product.}
\begin{align*}
  \lambda_{\mathsf{max}}=\max\; |\,\text{eig}(\bA\odot\bW)\,| < 1
\end{align*}
\citep[see][]{Daley-1988}. When we are conditioning on finite datasets we do not have to worry about this. We simply place weak priors on the network parameters, e.g., a beta prior on the sparsity~${\rho}$ of an Erd\H{o}s-Renyi graph, and a Jeffreys prior on the scale of the gamma weight distribution. For the generative model, however, we would like to set our hyperparameters such that the prior distribution places little mass on unstable networks. In order to do so, we use tools from random matrix theory. 

The celebrated circular law describes the asymptotic eigenvalue distribution for $K\times K$~random matrices with entries that are i.i.d. with zero mean and variance~$\sigma^2$. As~$K$ grows, the eigenvalues are uniformly distributed over a disk in the complex plane centered at the origin and with radius~$\sigma\sqrt{K}$. In our case, however, the mean of the entries,~${\mathbb{E}[A_{k,k'}W_{k,k'}] = \mu}$, is not zero.

\citet{Silverstein-1994} has shown that we can analyze noncentral random matrices by considering them to be perturbations about the mean. Consider~${\bA\odot\bW=\bV+\bU}$, where~${\bV=\mu K e_K e_K^T}$ is a deterministic rank-one matrix with every entry equal to~$\mu$, ${e_K\in\reals^K}$ is a column vector with all entries equal to~${K^{-1/2}}$,  and~$\bU$ is a random matrix with i.i.d.\ zero-mean entries. Then, as~$K$ approaches infinity, the largest eigenvalue will come from~$\bV$ and will be distributed as~${\lambda_{\sf{max}}\sim\distNormal(\mu K, \sigma^2)}$, and the remaining eigenvalues will be uniformly distributed over the complex disc.

\begin{sloppypar}
In the simple case of~${W_{k,k'}\sim \distGamma(\alpha,\beta)}$ and~${A_{k,k'}\sim\distBernoulli(\rho)}$, we have ${\mu = \rho \alpha/\beta}$ and ${\sigma=\sqrt{\rho((1-\rho)\alpha^2+\alpha)}/\beta}$. For a given~$K$,~$\alpha$ and~$\beta$, we can tune the sparsity parameter~$\rho$ to achieve stability with high probability. We simply set~$\rho$ such that the minimum of~$\sigma\sqrt{K}$ and, say,~${\mu K + 3\sigma}$, equals one. Figures~\ref{fig:stability_p_w} and~\ref{fig:stability_max_rho} show a variety of weight distributions and the maximum stable~$\rho$. Increasing the network size, the mean, or the variance will require a concomitant increase in sparsity. 
\end{sloppypar}

This approach relies on asymptotic eigenvalue distributions, and it is unclear how quickly the spectra of random matrices will converge to this distribution. To test this, we computed the empirical eigenvalue distribution for random matrices of various size, mean, and variance. We generated~$10^4$ random matrices for each weight distribution in Figure~\ref{fig:stability_p_w} with sizes~$K=4$,~$64$, and~$1024$, and~$\rho$ set to the theoretical maximum indicated by dots in Figure~\ref{fig:stability_max_rho}. The theoretical and empirical distributions of the maximum eigenvalue are shown in Figures~\ref{fig:stability_1_5} and~\ref{fig:stability_8_12}. We find that for small mean and variance weights, for example~$\distGamma(1,5)$ in the Figure~\ref{fig:stability_1_5}, the empirical results closely match the theory. As the weights grow larger, as in~${\distGamma(8,12)}$ in~\ref{fig:stability_8_12}, the empirical eigenvalue distributions have increased variance and lead to a greater than expected probability of unstable matrices for the range of network sizes tested here. We conclude that networks with strong weights should be counterbalanced by strong sparsity limits, or additional structure in the adjacency matrix that prohibits excitatory feedback loops.

\section{Synthetic Results}
\label{sec:synth}
Our inference algorithm is first tested on synthetic data generated from the network Hawkes model. We perform two tests: a) a link prediction task where the process identities are given and the goal is to simply infer whether or not an interaction exists, and b) an event prediction task where we measure the probability of held-out event sequences. 

The network Hawkes model can be used for link prediction by considering the posterior probability of interactions~$P(A_{k,k'}\given \{s_n,c_n\})$. By thresholding at varying probabilities we compute a ROC curve. A standard Hawkes process assumes a complete set of interactions~($A_{k,k'}\equiv 1$), but we can similarly threshold its inferred weight matrix to perform link prediction. 

Cross correlation provides a simple alternative measure of interaction. By summing the cross-correlation over offsets~${\Delta t\in[0,\Delta t_{\mathsf{max}})}$, we get a measure of directed interaction. A probabilistic alternative is offered by the generalized linear model for point processes (GLM), a popular model for spiking dynamics in computational neuroscience \cite{Paninski-2004}. The GLM allows for constant background rates and both excitatory and inhibitory interactions. Impulse responses are modeled with linear basis functions. Area under the impulse response provides a measure of directed excitatory interaction that we use to compute a ROC curve. See the supplementary material for a detailed description of this model.
\begin{figure}[t]
\begin{center}
\begin{subfigure}[b]{.49\linewidth}
\begin{center}
\includegraphics[width=.9\linewidth]{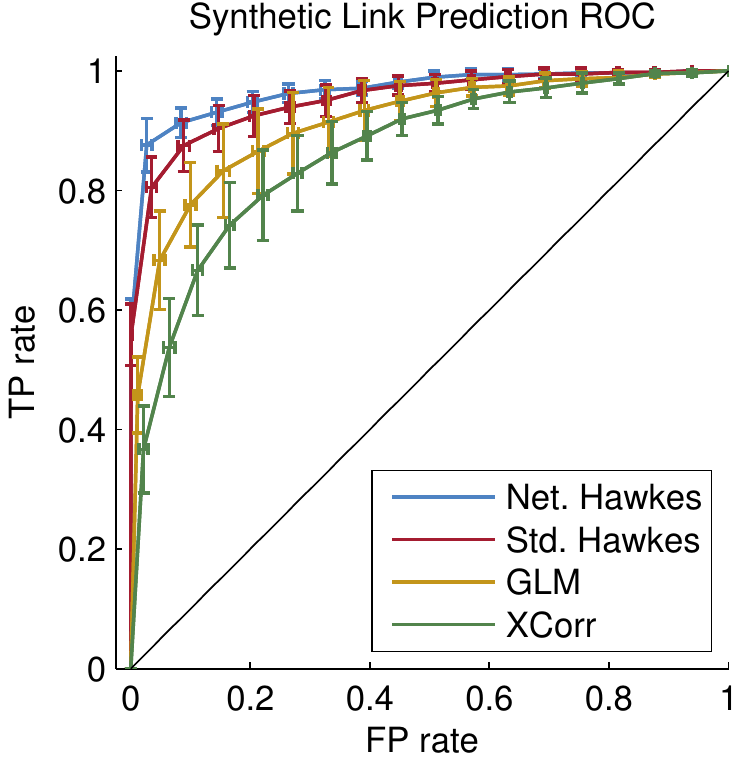} 
\end{center}
\caption{}
\label{fig:synth_link_pred}
\end{subfigure}
~
\begin{subfigure}[b]{.49\linewidth}
\begin{center}
\includegraphics[width=.9\linewidth]{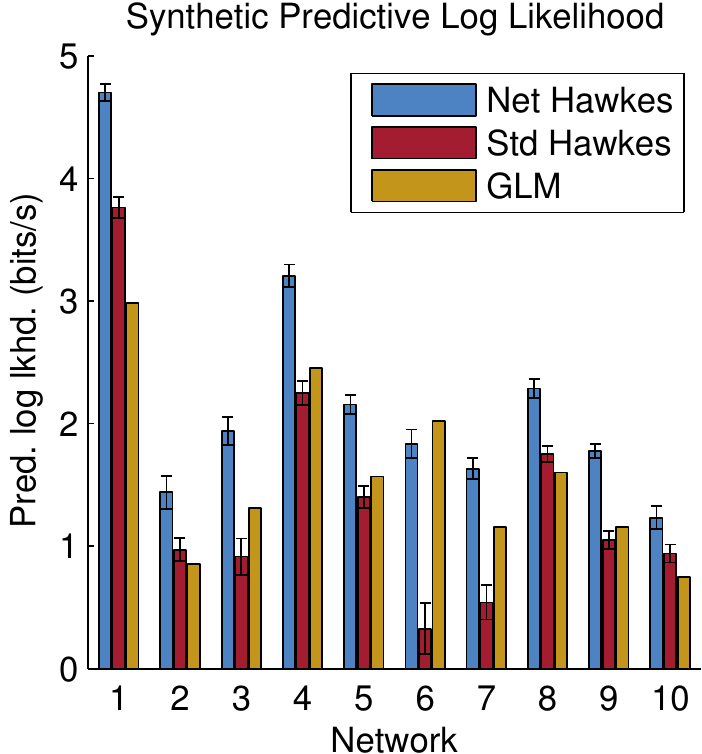} 
\end{center}
\caption{}
\label{fig:synth_pred_ll}
\end{subfigure}
\end{center}
\caption{(a) Comparison of models on a link prediction test averaged across ten randomly sampled synthetic networks of 30 nodes each. The network Hawkes model with the correct Erd\H{o}s-Renyi graph prior outperforms a standard Hawkes model, GLM, and simple thresholding of the cross-correlation matrix. (b) Comparison of predictive log likelihoods for the same set of networks as in Figure~\ref{fig:synth_link_pred}, compared to a baseline of a Poisson process with constant rate. Improvement in predictive likelihood over baseline is normalized by the number of events in the test data to obtain units of ``bits per spike.'' Again, the network Hawkes model outperforms the competitors in all but one sample network.}
\end{figure}

We sampled ten network Hawkes processes of~$30$ nodes each with Erd\H{o}s-Renyi graph models, constant background rates, and the priors described in Section~\ref{sec:basic_model}. The Hawkes processes were simulated for~${T=1000}$ seconds. We used the models above to predict the presence or absence of interactions. The results of this experiment are shown in the ROC curves of Figure~\ref{fig:synth_link_pred}. The network Hawkes model accurately identifies the sparse interactions, outperforming all other models.

With the Hawkes process and the GLM we can evaluate the log likelihood of held-out test data. On this task, the network Hawkes outperforms the competitors for 9 out 10 networks. On average, the network Hawkes model achieves~$2.2\pm.1$ bits/spike improvement in predictive log likelihood over a homogeneous Poisson process. Figure~\ref{fig:synth_pred_ll} shows that on average the standard Hawkes and the GLM provide only 60\% and 72\%, respectively, of this predictive power. See the supplementary material for further analysis.

\section{Trades on the S\&P 100}
\label{sec:financial}
As an example of how Hawkes processes may discover interpretable latent structure in real-world data, we study the trades on the S\&P~100 index collected at 1s intervals during the week of Sep.~28 through Oct.~2,~2009. Every time a stock price changes by~${\pm0.1\%}$ of its current price an event is logged on the stock's process, yielding a total of~${K=100}$ processes and~${N}$=182,037 events.

Trading volume varies substantially over the course of the day, with peaks at the opening and closing of the market. This daily variation is incorporated into the background rate via a Log Gaussian Cox Process (LGCP) with a periodic kernel (see supplementary material). We look for short-term interactions on top of this background rate with time scales of~${\Delta t_{\textsf{max}}=60\mathrm{s}}$. 

In Figure~\ref{fig:financial_pred_ll} we compare the predictive performance of independent LGCPs, a standard Hawkes process with LGCP background rates, and the network Hawkes model with LGCP background rates under two graph priors. The models are trained on four days of data and tested on the fifth. Though the network Hawkes is slightly outperformed by the standard Hawkes, the difference is small relative to the performance improvement from considering interactions, and the inferred network parameters provide interpretable insight into the market structure.
\begin{figure}[!h]
\vspace{1em}
\begin{subfigure}[T]{\linewidth}
\begin{center}
\begin{tabular}{|l|c|}
\hline
\textbf{Financial Model} & \textbf{Pred. log lkhd. (bits/spike)} \\
\hline
Indep. LGCP & $0.579\pm 0.006$ \\
Std. Hawkes & $0.903\pm 0.003$ \\
Net. Hawkes (Erd\H{o}s-Renyi) & $0.893\pm 0.003$ \\
Net. Hawkes (Latent Distance) & $0.879\pm 0.004$ \\
\hline
\end{tabular}
\end{center}
\end{subfigure}
\caption{Comparison of financial models on a event prediction task, relative to a homogeneous Poisson process baseline.}
\label{fig:financial_pred_ll}
\end{figure}

\begin{figure}[!t]
\begin{subfigure}[T]{\linewidth}
\begin{center}
\includegraphics[width=.55\linewidth]{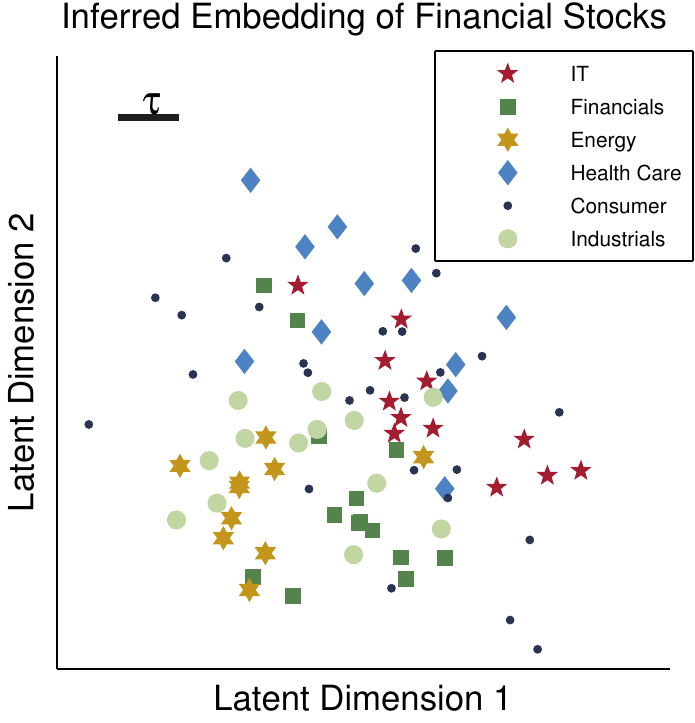} 
\end{center}
\end{subfigure}\\
\vskip1.5ex
\begin{subfigure}[T]{.8\linewidth}
\begin{center}
\includegraphics[width=\linewidth]{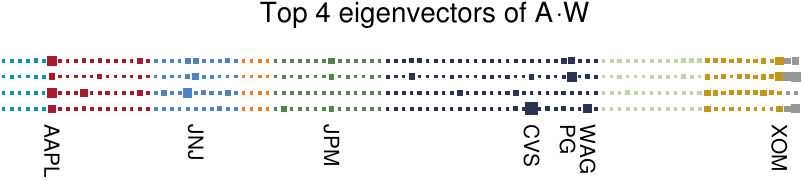} 
\end{center}
\end{subfigure}
\caption{Top: A sample from the posterior distribution over embeddings of stocks from the six largest sectors of the S\&P100 under a latent distance graph model with two latent dimensions. Scale bar: the characteristic length scale of the latent distance model. The latent embedding tends to embed stocks such that they are nearby to, and hence more likely to interact with, others in their sector. Bottom: Hinton diagram of the top 4 eigenvectors. Size indicates magnitude of each stock's component in the eigenvector and colors denote sectors as in the top panel, with the addition of Materials (aqua), Utilities (orange), and Telecomm (gray). We show the eigenvectors corresponding to the four largest eigenvalues ${\lambda_{\mathsf{max}}=0.74}$ (top row) to ${\lambda_4=0.34}$ (bottom row).}
\label{fig:financial_embedding}
\end{figure}

In the latent distance model for~$\bA$, each stock has a latent embedding~${\bx_k\in\reals^2}$ such that nearby stocks are more likely to interact, as described in Section~\ref{sec:graph_models}. Figure~\ref{fig:financial_embedding} shows a sample from the posterior distribution over embeddings in~$\reals^2$ for~${\rho=0.2}$ and~${\tau=1}$. We have plotted stocks in the six largest sectors, as listed on Bloomberg.com. Some sectors, notably energy and financials, tend to cluster together, indicating an increased probability of interaction between stocks in the same sector. Other sectors, such as consumer goods, are broadly distributed, suggesting that these stocks are less influenced by others in their sector. For the consumer industry, which is driven by slowly varying factors like inventory, this may not be surprising.  

The Hinton diagram in the bottom panel of Figure~\ref{fig:financial_embedding} shows the top 4 eigenvectors of the interaction network. All eigenvalues are less than 1, indicating that the system is stable. The top row corresponds to first eigenvector~(${\lambda_{\mathsf{max}}=0.74}$). Apple~(\texttt{AAPL}), J.P. Morgan~(\texttt{JPM}), and Exxon Mobil~(\texttt{XOM}) have notably large entries in the eigenvector, suggesting that their activity will spawn cascades of self-excitation. The fourth eigenvector~(${\lambda_4=0.34}$) is dominated by Walgreens~(\texttt{WAG}) and CVS~(\texttt{CVS}), suggesting bursts of activity in these drug stores, perhaps due to encouraging quarterly reports during flu season \cite{Walgreens-NYT-2009}.

\section{Gangs of Chicago}
\label{sec:chicago}
\begin{figure*}[!t]
\begin{center}
\begin{subfigure}[T]{.32\textwidth}
\begin{subfigure}[T]{\textwidth}
\begin{center}
\includegraphics[width=\linewidth]{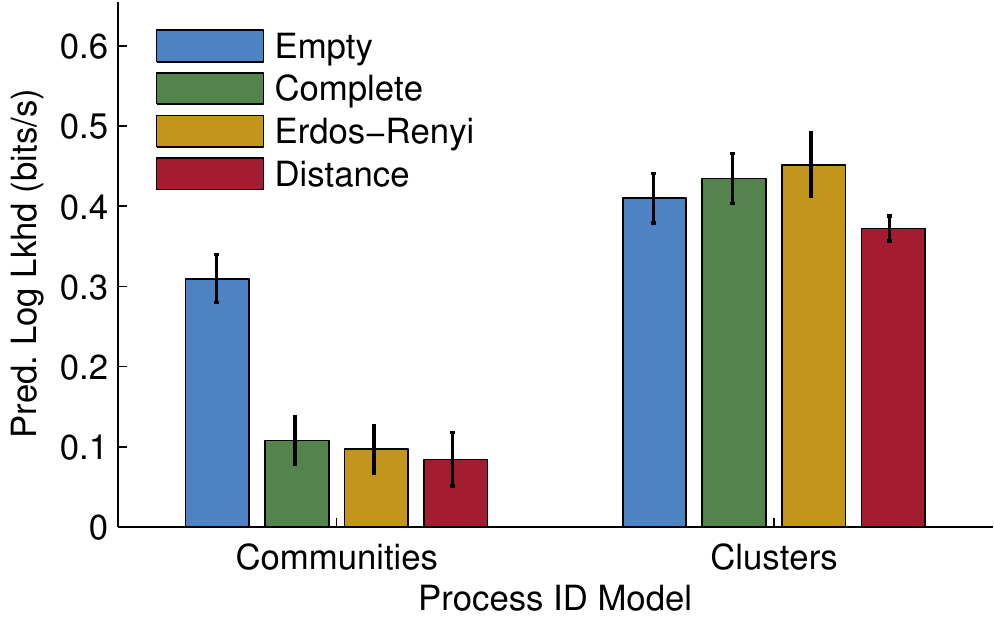}
\caption{}
\label{fig:chicago_predll}
\end{center}
\end{subfigure}
\begin{subfigure}[B]{\textwidth}
\vspace{1em}
\begin{center}
\includegraphics[width=.7\linewidth]{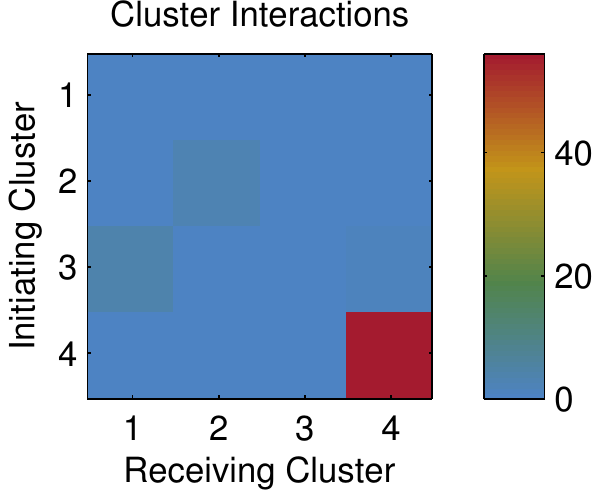}\\
\caption{}
\label{fig:chicago_interactions}
\end{center}
\end{subfigure}
\end{subfigure}
\begin{subfigure}[B]{.32\textwidth}
\includegraphics[width=\linewidth]{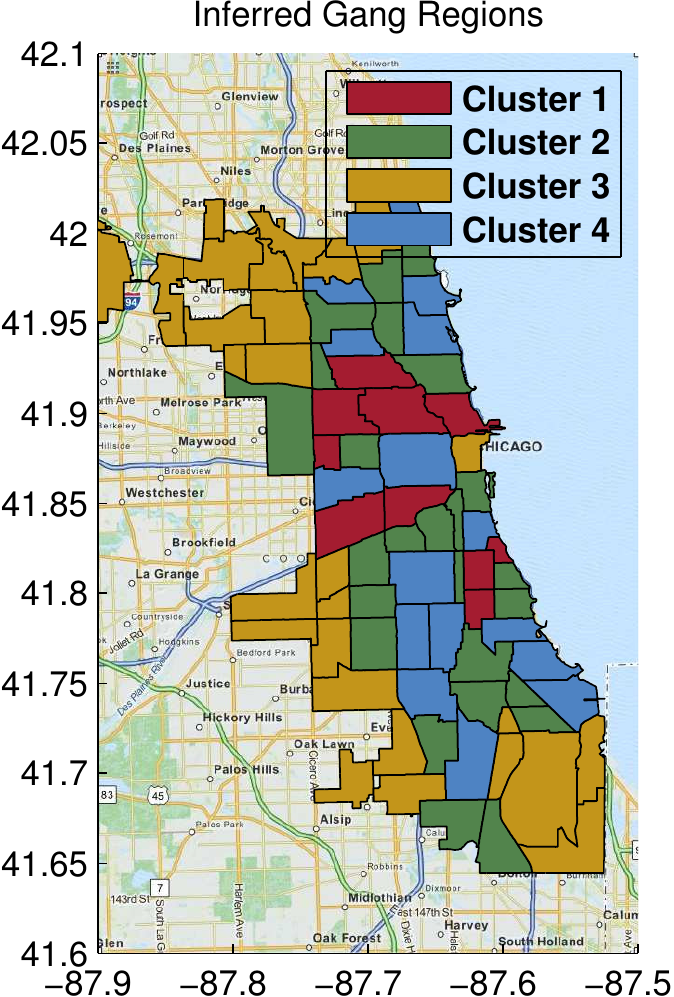} 
\caption{}
\label{fig:chicago_map}
\end{subfigure}
\begin{subfigure}[B]{.28\textwidth}
\includegraphics[width=\linewidth]{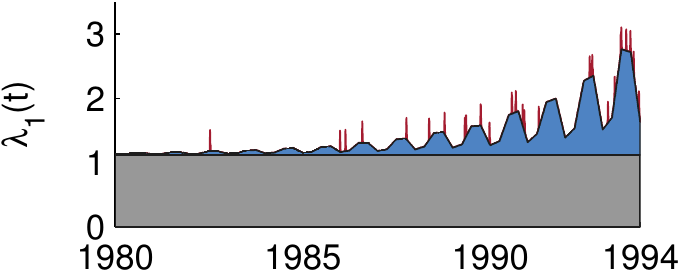} \\
\includegraphics[width=\linewidth]{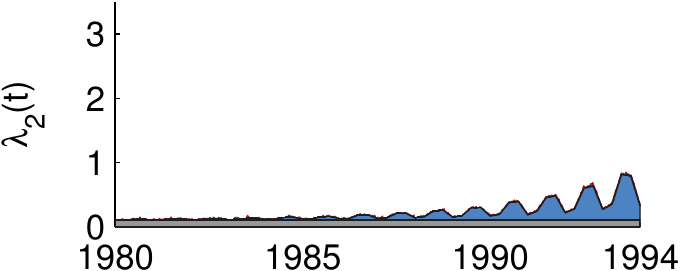} \\ 
\includegraphics[width=\linewidth]{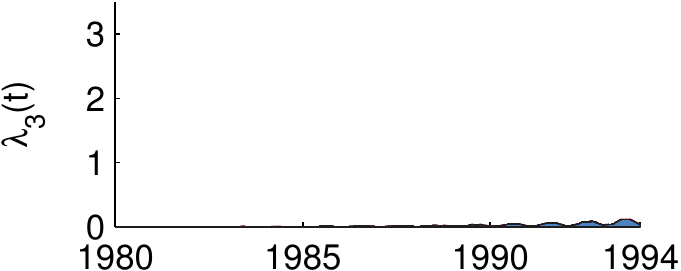} \\
\includegraphics[width=\linewidth]{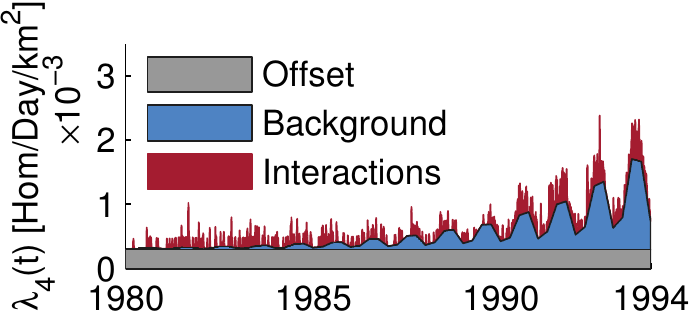} 
\caption{}
\label{fig:chicago_rates}
\end{subfigure}
\end{center}
\caption{Inferred interactions among clusters of community areas in the city of Chicago. (a) Predictive log likelihood for ``communities'' and ``clusters''  process identity models and four graph models. 
Panels (b-d) present results for the model with the highest predictive log likelihood: an Erd\H{o}s-Renyi graph with~${K=4}$ clusters.  (b) The weighted interaction network in units of induced homicides over the training period (1980-1993). (c) Inferred clustering of the 77 community areas. (d) The intensity for each cluster, broken down into the offset, the shared background rate, and the interactions (units of~${10^{-3}}$ homicides per day per square kilometer).}
\label{fig:chicago}
\end{figure*}

In our final example, we study spatiotemporal patterns of gang-related homicide in Chicago. Sociologists have suggested that gang-related homicide is mediated by underlying social networks and occurs in mutually-exciting, retaliatory patterns \cite{Papachristos-2009}. This is consistent with a spatiotemporal Hawkes process in which processes correspond to gang territories and homicides incite further homicides in rival territories.

We study gang-related homicides between 1980 and 1995 \cite{ICPSR}. Homicides are labeled by the community in which they occurred. Over this time-frame there were~${N=1637}$ gang-related homicides in the~${77}$ communities of Chicago. 

We evaluate our model with an event-prediction task, training on 1980-1993 and testing on 1994-1995. We use a Log Gaussian Cox Process (LGCP) temporal background rate in all model variations. Our baseline is a single process with a uniform spatial rate for the city.
We test two process identity models: a)~the ``community'' model, which considers each community a separate process, and b)~the ``cluster'' model, which groups communities into processes. The number of clusters is chosen by cross-validation (see supplementary material). For each process identity model, we compare four graph models: a)~independent LGCPs (\emph{empty}), b)~a standard Hawkes process with all possible interactions (\emph{complete}), c)~a network Hawkes model with a sparsity-inducing Erd\H{o}s-Renyi graph prior, and d)~a network Hawkes model with a latent distance model that prefers short-range interactions.

The community process identity model improves predictive performance by accounting for higher rates in South and West Chicago where gangs are deeply entrenched. Allowing for interactions between community areas, however, results in a decrease in predictive power due to overfitting (there is insufficient data to fit all~${77^2}$ potential interactions). Interestingly, sparse graph priors do not help. They bias the model toward sparser but stronger interactions which are not supported by the test data. These results are shown in the ``communities'' group of Figure~\ref{fig:chicago_predll}. Clustering the communities improves predictive performance for all graph models, as seen in the ``clusters'' group. Moreover, the clustered models benefit from the inclusion of excitatory interactions, with the highest predictive log likelihoods coming from a four-cluster Erd\H{o}s-Renyi graph model with interactions shown in Figure~\ref{fig:chicago_interactions}. Distance-dependent graph priors do not improve predictive performance on this dataset, suggesting that either interactions do not occur over short distances, or that local rivalries are not substantial enough to be discovered in our dataset. More data is necessary to conclusively say which.

Looking into the inferred clusters in Figure~\ref{fig:chicago_map} and their rates in~\ref{fig:chicago_rates}, we can interpret the clusters as ``safe suburbs'' in gold, ``buffer neighborhoods'' in green, and ``gang territories'' in red and blue. Self-excitation in the blue cluster (Figure~\ref{fig:chicago_interactions}) suggests that these regions are prone to bursts of activity, as one might expect during a turf-war. This interpretation is supported by reports of ``a burst of street-gang violence in 1990 and 1991'' in West Englewood (${41.77\degree}$N, ${-87.67\degree}$W) \cite{Block-1993}. 

Figure~\ref{fig:chicago_rates} also shows a significant increase in the homicide rate between 1989 and 1995, consistent with reports of escalating gang warfare \cite{Block-1993}. In addition to this long-term trend, homicide rates show a pronounced seasonal effect, peaking in the summer and tapering in the winter. A LGCP with a quadratic kernel point-wise added to a periodic kernel captures both effects.

\section{Related Work}
Multivariate point processes are of great interest to the machine learning community as they are intuitive models for a variety of natural phenomena. We have leveraged previous work on Poisson processes with Gaussian process intensities in our background rate models \cite{Cunningham-2008}. An expectation-maximization inference algorithm for Hawkes processes was put forth by \citet{Simma-2010} and applied to very large social network datasets. We have adapted their latent variable formulation in our fully-Bayesian inference algorithm and introduced a framework for prior distributions over the latent network. 

Others have considered special cases of the model we have proposed. \citet{Blundell-2012} combine Hawkes processes and the Infinite Relational Model (a specific exchangeable graph model with an Aldous-Hoover representation) to cluster  processes and discover interactions. \citet{Cho-2013} applied Hawkes processes to gang incidents in Los Angeles. They developed a spatial Gaussian mixture model (GMM) for process identities, but did not explore structured network priors. We experimented with this process identity model but found that it suffers in predictive log likelihood tests (see supplementary material). 

Recently, \citet{Iwata-2013} developed a stochastic EM algorithm for Hawkes processes, leveraging similar conjugacy properties, but without network priors. \citet{Zhou-2013} have developed a promising optimization-based approach to discovering low-rank networks in Hawkes processes, similar to some of the network models we explored.

Perhaps the most closely related work is that of \citet{Perry-2013}. They provide a partial likelihood inference algorithm for Hawkes processes with a similar emphasis on structural patterns in the network of interactions. They provide an estimator capable of discovering homophily (the tendency for similar processes to interact) and other network effects. Our fully-Bayesian approach generalizes this method to capitalize on recent developments in random network models \cite{Lloyd-2012} and allows for nonparametric background rates.

Finally, generalized linear models (GLMs) are widely used in computational neuroscience \cite{Paninski-2004}. GLMs allow for both excitatory and inhibitory interactions, but, as we have shown, when the data consists of purely excitatory interactions, Hawkes processes outperform GLMs in link- and event-prediction tests. 

\section{Conclusion}\label{discussion}


We developed a framework for discovering latent network structure from spiking data. Our auxiliary variable formulation of the multivariate Hawkes process supported arbitrary Aldous-Hoover graph priors, Log Gaussian Cox Process background rates, and models of unobserved process identities. 
Our parallel MCMC algorithm allowed us to reason about uncertainty in the latent network in a fully-Bayesian manner, taking into account noisy observations and prior beliefs. 
We leveraged results from random matrix theory to analyze the conditions under which random network models will be stable, and our applications uncovered interpretable latent networks in a variety of synthetic and real-world problems.
Generalizing beyond the Hawkes observation model is a promising avenue for future work.
\\
\\
\textbf{Acknowledgements.} The authors wish to thank Leslie Valiant for many valuable discussions. SWL is supported by a National Defense Science and Engineering Graduate Fellowship.
\clearpage
\bibliography{draft}
\bibliographystyle{icml2014}

\appendix

\section{Inference details}
\subsection{Derivation of conjugate prior updates}
By combining Equations~\ref{eq:poisson_lkhd}~and~\ref{eq:hawkes_likelihood} of the main text, we can write the joint likelihood, with the auxiliary parent variables, as,
\begin{multline*}\label{eq:complete_likelihood}
  p(\{s_n,c_n,z_n\}^N_{n=1},\given \{\lambda_{0,k}(t)\}^K_{k=1}, \{h_{k,k'}(\Delta t)\}_{k,k'}) = \\
  \prod^K_{k=1} \bigg[
  \exp\left\{ -\int_0^T \lambda_{0,k}(\tau)\mathrm{d}\tau \right\} \,
  \prod^N_{n=1}
   \lambda_{0,k}(s_n)^{\delta_{c_n,k}\delta_{z_n,0}} \bigg]\\
  \times \prod_{n=1}^N\prod_{k'=1}^K \bigg[
  \exp\left\{-\int^T_{s_n} h_{c_n,k'}(\tau - s_n)\mathrm{d}\tau \right\}\prod^N_{n'=1} h_{c_n,c_{n'}}(s_{n'}-s_n)^{\delta_{c_{n'},k'}\delta_{z_{n'},n}}\bigg].
\end{multline*}
The first line corresponds to the likelihood of the background processes; the second and third correspond to the likelihood of the induced processes triggered by each spike.

To derive the updates for weights, recall from Equation~\ref{eq:ir_decomp} of the main text that~${W_{k,k'}}$ only appears in the impulse responses for which~${c_n=k}$ and~${c_{n'}=k'}$. so we have,
\begin{align*}
p(W_{k,k'}\given& \{s_n,c_n,z_n\}^N_{n=1},\ldots) \\
&\propto\prod_{n=1}^N\left[ \exp\left\{-\int^T_{s_n} h_{k,k'}(\tau - s_n)\mathrm{d}\tau \right\}\prod_{n'=1}^N h_{k,k'}(s_{n'}-s_n)^{\delta_{c_{n'},k'}\delta_{z_{n'},n}}\right]^{\delta_{c_n,k}} \times p(W_{k,k'})\\
&=\prod_{n=1}^N\left[ \exp\left\{-\int^T_{s_n} A_{k,k'}W_{k,k'}g_{k,k'}(\tau - s_n)\mathrm{d}\tau \right\}\right.\\
&\qquad\qquad\left.\prod_{n'=1}^N \left(A_{k,k'}W_{k,k'}g_{k,k'}(s_{n'}-s_n)\right)^{\delta_{c_{n'},k'}\delta_{z_{n'},n}}\right]^{\delta_{c_n,k}}\times p(W_{k,k'}).
\end{align*}
If~${A_{k,k'}=1}$ and we ignore spikes after~${T-\Delta t_{\mathsf{max}}}$, this is approximately proportional to
\begin{align*}
&\exp\left\{-W_{k,k'}N_k\right\} W_{k,k'}^{N_{k,k'}} p(W_{k,k'}),
\end{align*}
where
\begin{align*}
N_{k}=\sum_{n=1}^N \delta_{c_n,k},\;\text{and}\;
N_{k,k'}=\sum_{n=1}^N\sum_{n'=1}^N \delta_{c_n,k}\delta_{c_{n'},k'}\delta_{z_{n'},n}.
\end{align*}
When~${p(W_{k,k'})}$ is a gamma distribution, the conditional distribution is also gamma. If~${A_{k,k'}=0}$, the conditional distribution reduces to the prior, as expected.

Similar conjugate updates can be derived for constant background rates and the impulse response parameters, as stated in the main text.

\subsection{Log Gaussian Cox Process background rates}
In the Trades on the S\&P100 and the Gangs of Chicago datasets, it was crucial to model the background fluctuations that were shared among all processes. However, if the background rate is allowed to vary at time scales shorter than~${\Delta t_{\textsf{max}}}$ then it may obscure interactions between processes. To prevent this, we sample the Log Gaussian Cox Process (LGCP) at a sparse grid of~$M+1$ equally spaced points and linearly interpolate to evaluate the background rate at the exact time of each event. We have,
\begin{align*}
\by=\left\{\hat{y}\left(\frac{mT}{M}\right)\right\}_{m=0}^M \sim \mathcal{GP}(\boldsymbol{0},K(t,t')).
\end{align*}
Then,

\begin{align*}
\left\{\hat{\lambda}_{0,k}\left(\frac{mT}{M}\right)\right\}_{m=0}^M &= \mu_k + \alpha_k\exp\left\{\hat{y}\left(\frac{mT}{M}\right)\right\},
\end{align*}
and~${\lambda_{0,k}(s_n)}$ is linearly interpolated between the rate at surrounding grid points.

The equally spaced grid allows us to calculate the integral using the trapezoid quadrature rule. We use Elliptical Slice Sampling \cite{Murray-2010} to sample the conditional distribution of the vector ~$\by$.

Kernel parameters are set empirically or with prior knowledge. For example, the period of the kernel is set to one day for the S\&P100 dataset and one year for the Gangs of Chicago dataset since these are well-known trends. The scale and offset parameters have log Normal priors set such that the maximum and minimum homogeneous event counts in the training data are within two standard deviations of the expected value under the LGCP background rate. That is, the background rate should be able to explain all of the data without any observations if there is no evidence for interactions.

\subsection{Priors on hyperparameters}
When possible, we sample the parameters of the prior distributions. For example, in the Erd\H{o}s-Renyi graph model we place a~${\mathrm{Beta}(1,1)}$ prior on the sparsity~$\rho$. For the latent distance model, we place a log normal prior on the characteristic length scale~${\tau}$ and sample it using Hamiltonian Monte Carlo. 

For all of the results in this paper, we fixed the prior on the interaction kernel,${~g(\Delta t)}$ to a weak Normal-Gamma distribution with parameters ${\mu_\mu^0=-1.0}$, ${\kappa_\mu^0=10}$, ${\alpha_\tau^0=10}$, and ${\beta_\tau^0=1}$.

\paragraph{Scale of gamma prior on weights.}
For real data, we place an uninformative prior on the weight distribution. The gamma distribution is parameterized by a shape $\alpha_W^0$ and an inverse scale or rate $\beta_W^{0}$. The shape parameter~${\alpha_W^0}$ is chosen by hand (typically we use~${\alpha_W^0=2}$), but the inverse scale parameter~${\beta_W^0}$ is sampled. We may not know a proper scale a priori, however we can use a scale-invariant Jeffrey's prior to infer this parameter as well. Jeffrey's prior is proportional to the square root of the Fisher information, which for the gamma distribution is
\begin{align*}
  \Pr(\beta_W^0) \propto \sqrt{I(\beta_W^0)} = \frac{\sqrt{\alpha_W^0}}{\beta_W^0}.
\end{align*}
Hence the posterior is 
\begin{align*}
\Pr(\beta_W^0 \given \{\{W_{k,k'}\}\} ) &\propto \frac{\sqrt{\alpha_W^0}}{\beta_W^0}\prod_{k=1}^K\prod_{k'=1}^K \frac{(\beta_W^0)^{\alpha_W^0}}{\Gamma(\alpha_W^0)} W_{k,k'}^{\alpha_W^0-1}e^{-\beta_W^0 W_{k,k'}} \\
&\propto (\beta_W^0)^{K^2\alpha_W^0-1} \exp\left\{-\beta_W^0 \sum_{k=1}^{K}\sum_{k'=1}^K W_{k,k'}\right\}.
\end{align*}
This is a gamma distribution with parameters,
\begin{align*}
\beta_W^0\sim\distGamma(K^2\alpha_W^0,\sum_{k=1}^K\sum_{k'=1}^K W_{k,k'}).
\end{align*}

\section{Synthetic test details}
We generated~${T=1000}$s of events for each synthetic network. The average number of spikes was 25,732~$\pm$~9,425. Network 6, the only network for which the GLM outperformed the network Hawkes model in the event-prediction test,  was an outlier with 44,973 events. For event prediction, we trained on the first 900 seconds and tested on the last 100 seconds of the data. We ran our Markov chain for 2500 iterations and computed the posterior probabilities of~$\bA$ and~$\bW$ using the last 500 samples. 

A simple alternative to the Hawkes model is to look at cross-correlation between the event times. First, the event times are binned into an array~$\hat{\bs}_k$ of length~$M$. Let~${(\hat{\bs}_k \star \hat{\bs}_{k'})[m]}$ be the cross-correlation between~$\hat{\bs}_k$ and~$\hat{\bs}_{k'}$ at discrete time lag~$m$. Then,~${W_{k,k'}=\sum_{m=0}^{\Delta t_{\mathsf{max}}M/T}(\hat{\bs}_k \star \hat{\bs}_{k'})[m]}$ provides a simple measure of directed, excitatory interaction that can be thresholded to perform link prediction.

Additionally, we compare the network Hawkes process to the generalized linear model for point processes, a popular model in computational neuroscience \cite{Paninski-2004}. Here, the event counts are modeled as ${\hat{s}_{k,m}\sim\text{Poisson}(\lambda_{k,m})}$. The mean depends on external covariates and other events according to
\begin{align*}
  \lambda_{k,m} &= \exp\left\{ \balpha_k^T\by_m + \sum_{k'=1}^K\sum_{b=1}^B \beta_{k,k',b} (g_{b}\ast \hat{s}_{k'})[m ]\right\},
\end{align*}
where $\by_m$ is an external covariate at time~$m$, $\{g_b(\Delta m)\}_{b=1}^B$ are a set of basis functions that model impulse responses, and $\balpha$ and $\bbeta$ are parameters to be inferred. Under this formulation the log-likelihood of the events is concave function of the parameters and is easily maximized. Unlike the Hawkes process, however, this model allows for inhibitory interactions. 

For link prediction,~${\sum_b \beta_{k,k',b}}$ provides a measure of directed excitatory interaction that can be used to compute an ROC curve. In our comparisons, we used~${\by_m\equiv 1}$ to allow for time-homogeneous background activity and set~${\{g_b(\Delta m)\}}$ to the top~${B=6}$ principal components of a set of logistic normal impulse responses randomly sampled from the Hawkes prior. 

We used an L1 penalty to promote sparsity in the parameters of the GLM, and chosen the penalty using cross validation on the last 100 seconds of the training data.

\begin{figure}[h!]
\begin{center}
\begin{subfigure}[T]{\linewidth}
\begin{center}
\begin{tabular}{|l|c|}
\hline
\textbf{Model} & \textbf{Relative prediction improvement} \\
\hline
Network Hawkes & 100\% \\
Standard Hawkes & 59.2$\pm$14.2\% \\
GLM & 71.6$\pm$9.2\%\\
\hline
\end{tabular}
\end{center}
\end{subfigure}
\end{center}
\caption{Relative improvement in predictive log likelihood over a homogeneous Poisson process baseline. Relative to the network Hawkes, the standard Hawkes and the GLM yield significantly less predictive power.}
\label{tab:rel_pred_ll}
\end{figure}
Figure~\ref{fig:synth_pred_ll} of the main text shows the predictive log likelihoods for the Hawkes model with the correct Erd\"os-Renyi prior, the standard Hawkes model with a complete graph of interactions, and a GLM. On all but network 6, the network Hawkes model outperforms the competing models in terms of predictive log likelihood. Table~\ref{tab:rel_pred_ll} shows the average predictive performance across sample nextworks. The standard Hawkes and the GLM provide only 59.2\% and 71.6\%, respectively, of this predictive power.

\section{Trades on the S\&P100 model details}
We study the trades on the S\&P~100 index collected at 1s intervals during the week of Sep.~28 through Oct.~2,~2009. We group both positive and negative changes in price into the same process in order to measure overall activity. Another alternative would be to generate an ``uptick'' and a ``downtick'' process for each stock.  We ignored trades outside regular trading hours because they tend to be outliers with widely varying prices. Since we are interested in short term interactions,  we chose~${\Delta t_{\mathsf{max}}=60\mathrm{s}}$. This also limits the number of potential event parents. If we were interested in interactions over longer durations, we would have to threshold the price changes at a higher level. We precluded self-excitation for this dataset since upticks are often followed by downticks and vice-versa. We are seeking to explain these brief price jumps using the activity of other stocks.

We run our Markov chain for 2000 iterations and compute predictive log likelihoods and the eigenvalues of the expected interaction matrix,~${\mathbb{E}[\bA\odot\bW]}$, using the last 400 iterations of the chain. The posterior sample illustrated in the main text is the last sample of the chain.

\begin{figure}[!b]
\begin{subfigure}[T]{\linewidth}
\begin{center}
\includegraphics[width=.5\linewidth]{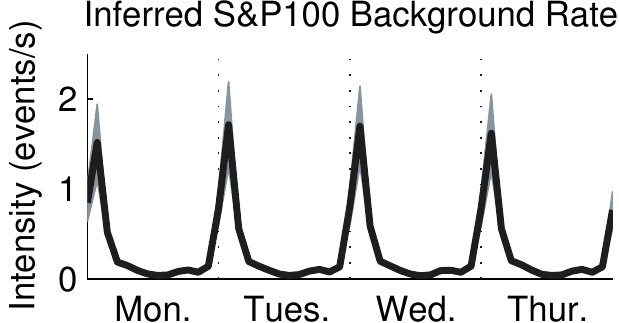}
\end{center}
\end{subfigure}
\caption{Posterior distribution over shared background rates for the S\&P100. Shading indicates two standard deviations from the mean.}
\label{fig:financial_bkgd}
\end{figure}

Trading volume varies substantially over the course of the day, with peaks at the opening and closing of the market. This daily variation is incorporated into the background rate via a Log Gaussian Cox Process with a periodic kernel. We set the period to one day. Figure~\ref{fig:financial_bkgd} shows the posterior distribution over the background rate.

Though it is not discussed in the main text, we also considered stochastic block model (SBM) priors as well \cite{Hoff-2008}, in hopes of recovering latent sector affiliations based on patterns of interaction between sectors. For example, stocks in the financial sector may have 90\% probability of interacting with one another, and 30\% probability of interacting with stocks in the energy sector. Rather than trying to interpret this from the embedding of a latent distance model, we can capture this belief explicitly with a stochastic block model prior on connectivity. We suppose there are~$J$ sectors, and the probability of belonging to a given sector is~${\balpha \in [0,1]^J\sim\text{Dirichlet}(\balpha_0)}$. The latent sector assignments are represented by the vector~${\bb\in [1,J]^K}$, where~${b_k\sim\text{Cat}(\balpha)}$. The probability of a directed interaction is~${\Pr(A_{k,k'}=1)=B_{b_{k},b_{k'}}}$, where~$\bB$ is a~${J\times J}$ matrix of Bernoulli probabilities. We place a beta prior on the entries of $\bB$. 

Our experiments with the SBM prior yield comparable predictive performance to the latent distance prior, as shown in Figure~\ref{fig:financial_pred_ll_sbm}. The inferred clusters (not shown) are correlated with the clusters identified by Bloomberg.com, but more analysis is needed. It would also be interesting to study the difference in inferred interactions under the various graph models; this is left for future work.

\begin{figure}[!h]
\begin{subfigure}[T]{\linewidth}
\begin{center}
\begin{tabular}{|l|c|}
\hline
\textbf{Financial Model} & \textbf{Pred. log lkhd. (bits/spike)} \\
\hline
Indep. LGCP & $0.579\pm 0.006$ \\
Std. Hawkes & $0.903\pm 0.003$ \\
Net. Hawkes (Erd\H{o}s-Renyi) & $0.893\pm 0.003$ \\
Net. Hawkes (Latent Distance) & $0.879\pm 0.004$ \\
Net. Hawkes (SBM) & $0.882\pm 0.004$ \\
\hline
\end{tabular}
\end{center}
\end{subfigure}
\caption{Comparison of financial models on a event prediction task, relative to a homogeneous Poisson process baseline.}
\label{fig:financial_pred_ll_sbm}
\end{figure}

\section{Gangs of Chicago model details}

\begin{figure*}[!b]
\begin{subfigure}[T]{.24\linewidth}
\begin{center}
\includegraphics[width=\linewidth]{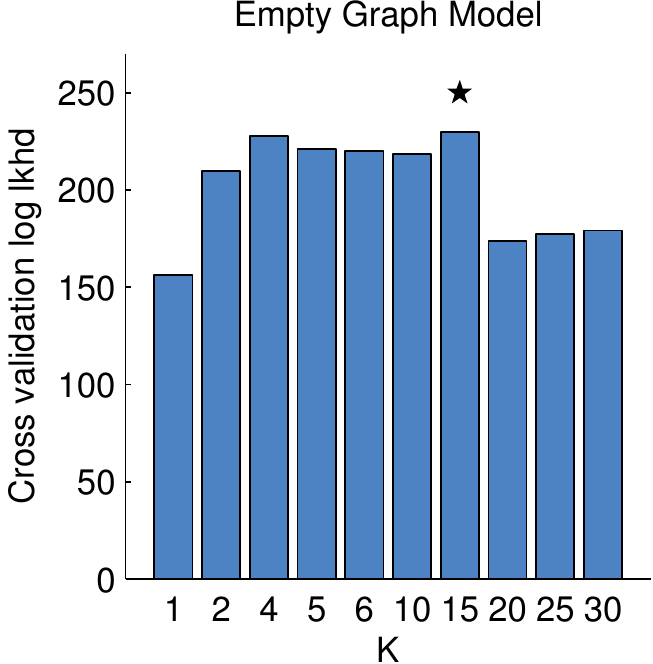}
\end{center}
\end{subfigure}
\begin{subfigure}[T]{.24\linewidth}
\begin{center}
\includegraphics[width=\linewidth]{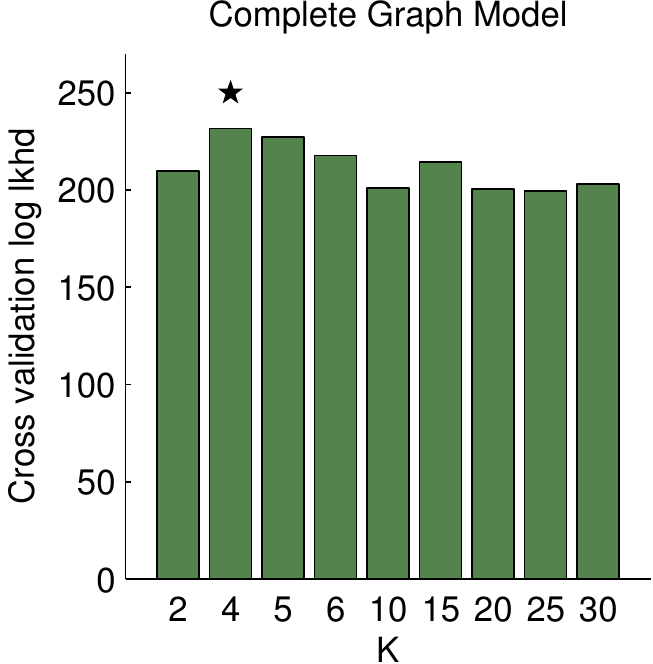}
\end{center}
\end{subfigure}
\begin{subfigure}[T]{.24\linewidth}
\begin{center}
\includegraphics[width=\linewidth]{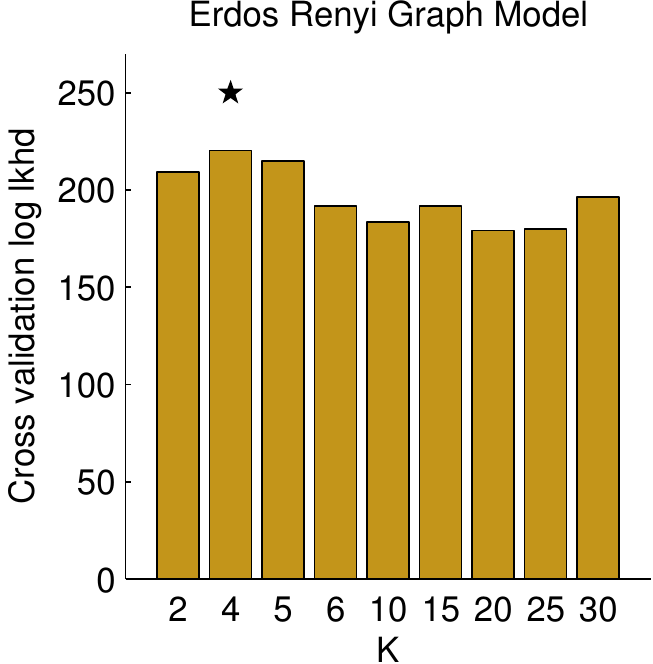}
\end{center}
\end{subfigure}
\begin{subfigure}[T]{.24\linewidth}
\begin{center}
\includegraphics[width=\linewidth]{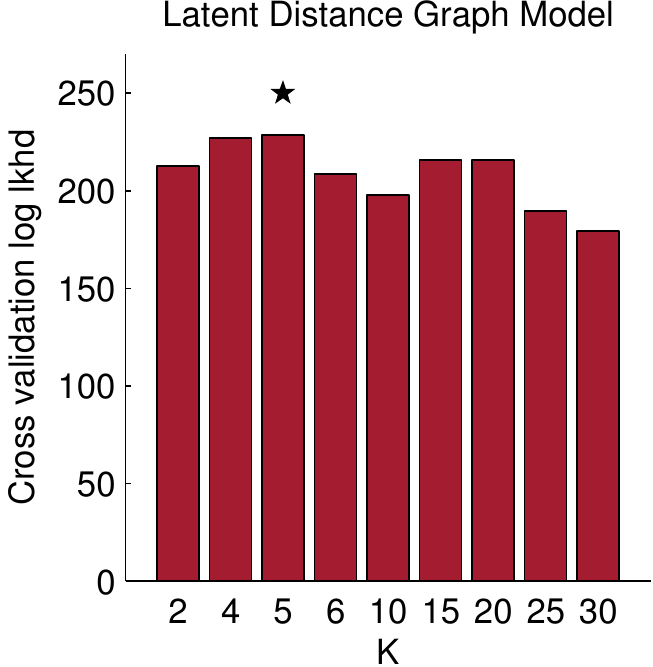}
\end{center}
\end{subfigure}
\caption{Cross validation results for Chicago models with~$K$ clusters for each of the four graph models.}
\label{fig:chicago_xv}
\end{figure*}
The first 12 years are used for training, 1993 is reserved for cross-validation, and the remaining two years are used to test the predictive power of the models. We also considered the crime dataset from \url{www.data.cityofchicago.org}, but this does not identify gang-related incidents.

We run our Markov chain for 700 iterations and use the last 200 iterations to compute predictive likelihoods and expectations. The posterior sample illustrated in the figure in main text is the last sample of the chain.
Since this is a spatiotemporal dataset, our intensities are functions of both spatial location and time. For simplicity we factorize the intensity into~${\lambda_{k,x}(\bx)\lambda_{k,t}(t)}$, where~${\lambda_{k,t}(t)}$ is a Gaussian process as described above, and~${\lambda_{k,x}(\bx)}$ is uniformly distributed over the spatial region associated with process~$k$ and is normalized such that it integrates to~$1$. 

In the case of the latent distance model with the community process model, each community's location is fixed to its center of mass. With the cluster process model, we introduce a latent location for each cluster and use a Gaussian distribution for the prior probability that a community belongs to a cluster. This encourages spatially localized clusters.

Figure~\ref{fig:chicago_xv} shows the cross validation results used to select the number of clusters,~$K$, in the clustered process identity model and each of the four graph models. For the empty, complete, and Erd\"os-Renyi graph priors, we discover~${K=15}$, 4, and 4 clusters respectively. The latent distance model, with its prior for spatially localized clusters, has its best performance for~${K=5}$ clusters. 

\begin{figure}[!t]
\begin{center}
\begin{subfigure}[T]{.5\linewidth}
\includegraphics[width=\linewidth]{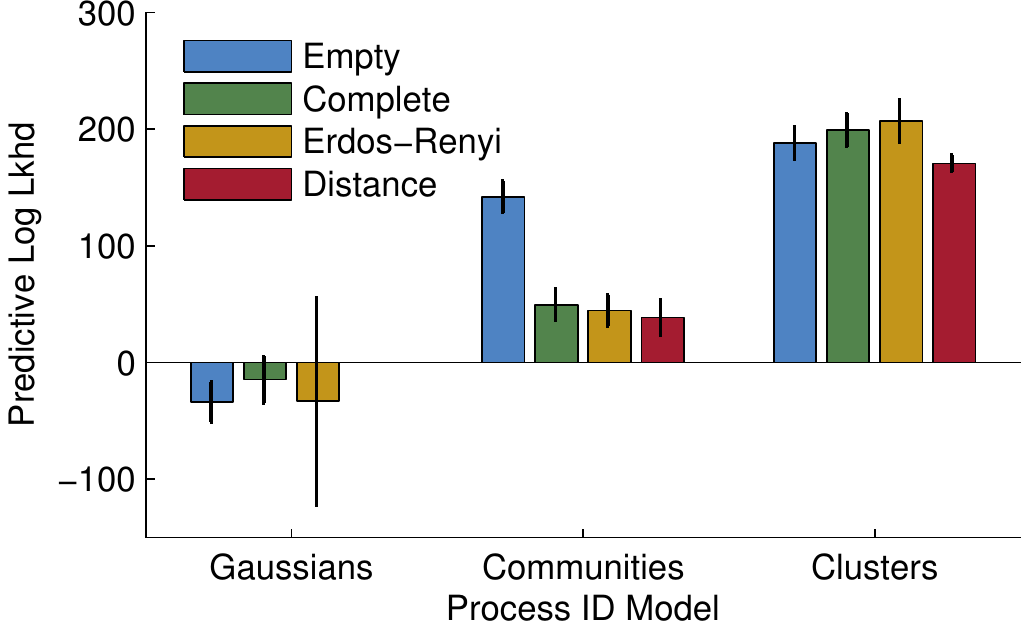}
\end{subfigure}
\end{center}
\caption{Comparison of predictive log likelihoods for Chicago homicides. This is the same as Figure~\ref{fig:chicago_predll} of the main text, but also includes the spatial GMM process identity model.}
\label{fig:chicago_pred_ll_gaussian}
\end{figure}

The spatial GMM process ID model from \citet{Cho-2013} fails on this dataset because it assigns its spatial intensity over all of~$\reals^2$, whereas the clustering model concentrates the rate on only the communities in which the data resides. Figure~\ref{fig:chicago_pred_ll_gaussian} shows the results of this spatial process ID model on the prediction task. We did not test a latent distance model with the spatial GMM, but it would likely suffer in the same way as the empty, complete, and Erd\H{o}s-Renyi graph priors.

\end{document}